\documentclass[11pt, a4paper, logo, copyright, nonumbering]{klear}
\usepackage[authoryear, compress, round]{natbib}
\usepackage{dblfloatfix}
\usepackage{caption}
\usepackage{dramatist}
\usepackage{xspace}
\usepackage{pifont} 
\usepackage{multirow}
\usepackage{tcolorbox}
\usepackage{xltabular}
\usepackage{longtable}
\usepackage{hyperref}
\usepackage{amsfonts}
\usepackage{amsmath}
\usepackage{amssymb}
\usepackage{lineno}
\usepackage{multirow}
\usepackage{adjustbox}

\usepackage[bottom]{footmisc}

\usepackage{CJKutf8}
\usepackage{subfigure}
\usepackage{setspace}

\usepackage{dsfont}
\usepackage{array} %
\usepackage{tabularx} %
\usepackage{subfigure} %
\usepackage{xcolor} %
\usepackage{algorithm}
\usepackage{algorithmic}

\usepackage{lipsum}  %
\usepackage{multicol} %

\usepackage{graphicx}
\usepackage{booktabs}
\usepackage{multirow}

\definecolor{keywordcolor}{rgb}{0.7, 0.1, 0.1}   %
\definecolor{tacticcolor}{rgb}{0.0, 0.1, 0.6}    %
\definecolor{commentcolor}{rgb}{0.4, 0.4, 0.4}   %
\definecolor{symbolcolor}{rgb}{0.0, 0.1, 0.6}    %
\definecolor{sortcolor}{rgb}{0.1, 0.5, 0.1}      %
\definecolor{attributecolor}{rgb}{0.7, 0.1, 0.1} %

\usepackage{listings}
\usepackage{textcomp}
\usepackage{minted}
\usepackage{subcaption}

\usepackage{tcolorbox}
\newtcolorbox{promptbox}[1][]{
    colback=gray!10,      %
    colframe=black!50,     %
    boxrule=0.5mm,        %
    arc=1mm,              %
    boxsep=0mm,
    fontupper=\ttfamily\scriptsize,  %
    width=\textwidth,     %
    title=#1,
fonttitle=\ttfamily\footnotesize\centering,
}

\makeatletter
\def\@BTrule[#1]{%
  \ifx\longtable\undefined
    \let\@BTswitch\@BTnormal
  \else\ifx\hline\LT@hline
    \nobreak
    \let\@BTswitch\@BLTrule
  \else
     \let\@BTswitch\@BTnormal
  \fi\fi
  \global\@thisrulewidth=#1\relax
  \ifnum\@thisruleclass=\tw@\vskip\@aboverulesep\else
  \ifnum\@lastruleclass=\z@\vskip\@aboverulesep\else
  \ifnum\@lastruleclass=\@ne\vskip\doublerulesep\fi\fi\fi
  \@BTswitch}
\makeatother

\addto\extrasenglish{
}

 {\begin{list}{}%
         {\setlength{\leftmargin}{#1}}%
         \item[]%
 }
 {\end{list}}
 
\reportnumber{001} %

\renewcommand{\today}{}

{\centering
\title{Klear-AgentForge: Forging Agentic Intelligence through Posttraining Scaling}}

\author[*]{

Klear Team, Kuaishou Technology
}

\begin{abstract}
    Despite the proliferation of powerful agentic models, the lack of critical post-training details hinders the development of strong counterparts in the open-source community. In this study, we present a comprehensive and fully open-source pipeline for training a high-performance agentic model for interacting with external tools and environments, named Klear-Qwen3-AgentForge, starting from the Qwen3-8B base model. We design effective supervised fine-tuning (SFT) with synthetic data followed by multi-turn reinforcement learning (RL) to unlock the potential for multiple diverse agentic tasks. We perform exclusive experiments on various agentic benchmarks in both tool use and coding domains. 
    Klear-Qwen3-AgentForge-8B achieves state-of-the-art performance among LLMs of similar size and remains competitive with significantly larger models. %
\end{abstract}

\begin{document}
\begin{CJK*}{UTF8}{gbsn}

\maketitle

\begin{figure}[h]
\vspace{-1em}
\centering
    \begin{minipage}[t]{0.98\columnwidth}
        \centering
        \includegraphics[width=\textwidth]{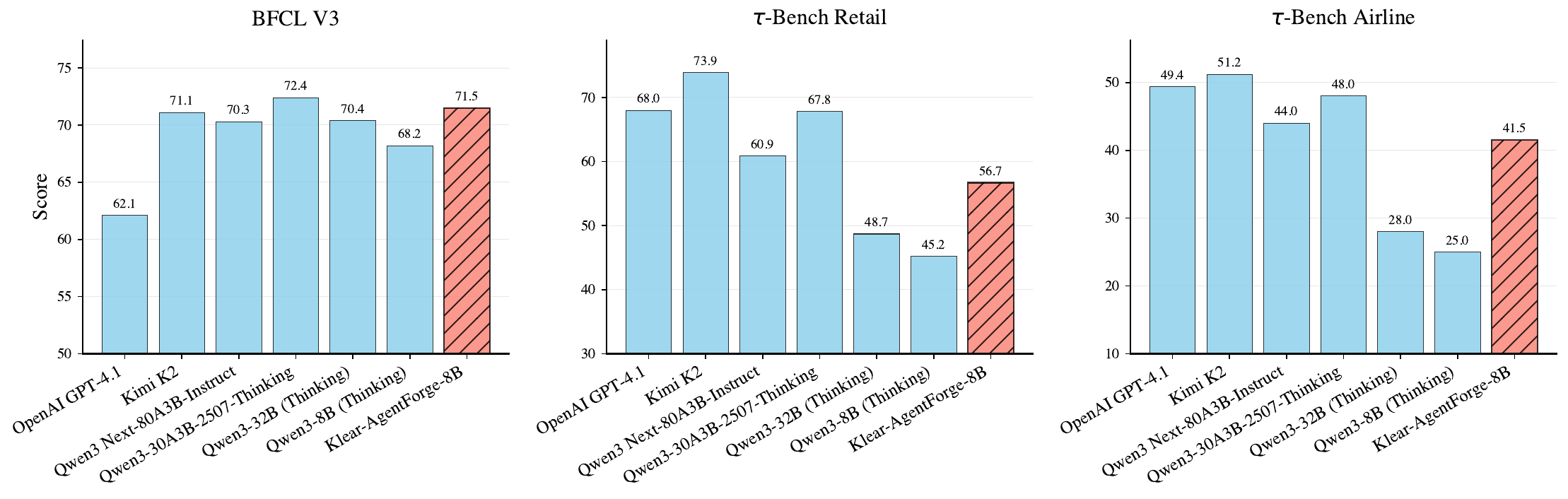}
    \end{minipage}
    \hfill 
    \begin{minipage}[t]{0.98\columnwidth}
        \centering
        \includegraphics[width=\textwidth]{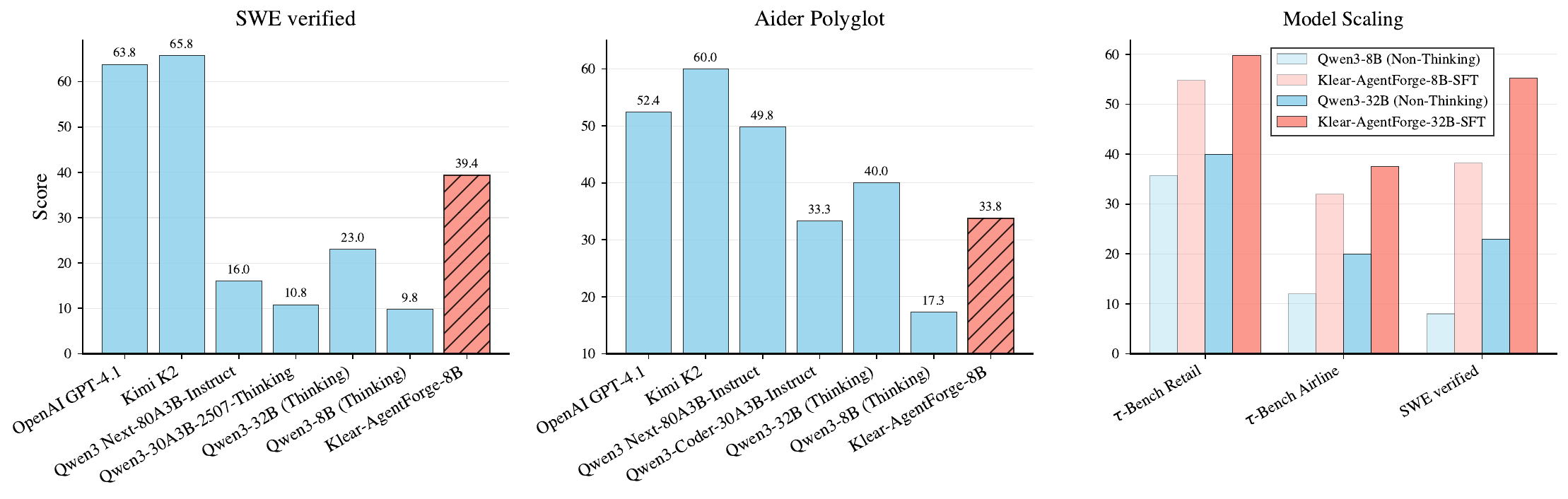}
    \end{minipage}
    \caption{
        Klear-AgentForge main results on multiple agentic tasks. 
    }
\label{fig:RL_performance}
\vspace{-2em}
\end{figure}

\newpage
\tableofcontents
\newpage

\section{Introduction}
\label{sec:intro}

Recent advances have highlighted the growing importance of the agentic paradigm, where models are not limited to single-step responses but instead act autonomously over multiple iterations to achieve complex goals~\citep{claudesonnet-4,gpt4.1,yang2025qwen3,team2025kimi,zeng2025glm}. In domains such as coding, this paradigm is particularly valuable: solving programming problems often requires decomposition, planning, execution, and verification across multiple reasoning cycles. Agentic sytems enable models to iteratively refine their thoughts, interact with environments or tools, and self-correct based on intermediate feedback. This multi-turn process allows for deeper problem solving and more reliable outcomes compared to traditional single-pass generation.
However, the transition from a next-token-prediction large language model (LLM) to a versatile, agentic LLM capable of directly constructing solutions for a wide range of agent tasks presents significant technical challenges. 

The primary challenge is that traditional LLMs must learn agentic behaviors through interactions in various environments, a process that is inherently inefficient and difficult to supervise. Several approaches have been proposed to scale up the generation of synthetic data for specific agents (\textit{e.g.}, coding~\citep{claudecode,cursor}, search~\citep{su2025scaling,2025mirothinker}, or tool-use~\citep{fang2025towards,prabhakar2025apigen,zhao2025muarlmultiturnuserinteractingagent}) and then employ specialized post-training. Despite these efforts, a comprehensive and effective training methodology remains elusive.We still lack a unified recipe for collecting high-quality trajectories and learning from them robustly.

The training paradigm also differs significantly, a distinction that lies mainly in the reinforcement learning (RL) framework.  In popularly studied reasoning tasks like math problem solving, RL focuses on single-turn decision-making, where an LLM receives immediate feedback after each rollout~\citep{deepseekai2025deepseekr1incentivizingreasoningcapability}.  In contrast, agentic RL extends this paradigm to multi-step reasoning processes. This means the agent must plan and act over longer horizons, often engaging in sequences of intermediate reasoning steps and environment interactions before reaching a final outcome. As a result, the reward signal in agentic RL tends to be much sparser—since meaningful feedback is only available at the end of a reasoning trajectory rather than after every individual step. This sparse reward setting introduces significant challenges for credit assignment and policy optimization, requiring more sophisticated exploration strategies and long-term value estimation compared to standard reasoning RL~\citep{tan2025processsupervisedreinforcementlearninginteractive,zeng2025glm}.

To address these challenges, we explore two complementary directions.
First, we develop systematic methods for constructing multi-domain, multi-task environments and synthesizing trajectories for agentic supervised learning. In this work, we focus on two domains: tool-use and coding. The coding domain encompasses both code contests' problems and software engineering tasks.
Second, we design a reinforcement learning framework featuring a fine-grained reward mechanism that provides localized feedback at intermediate actions to mitigate reward sparsity.
Together, these contributions form a complete training recipe to enhance the stability and efficiency of building agentic LLMs.

Using a two-stage training of supervised finetuning (SFT) and RL, we built Klear-AgentForge-8B from Qwen3-8B base~\citep{yang2025qwen3}. The resulting unified agentic LLM delivers state-of-the-art results on four benchmarks for tool-use and coding domains. We further present a granular analysis quantifying the effect of scaling SFT and RL compute on the model's final capabilities.

\section{Training Recipe}
\label{sec:method}
\subsection{Overview}
\label{subsec:overview}
\subsubsection{Agentic Task Formalization}
In agentic task, a complete rollout is a multi-turn interaction between the LLM and the environment. At each step, the LLM can choose:
\begin{itemize}
    \item {\textbf{Generation Action}}: Directly generate text (thinking, answering, summarizing, \textit{etc}.)
    \item {\textbf{Tool Call Action}}: Invoke external tools to gather information
\end{itemize}
A complete rollout trajectory follows the ReAct~\citep{yao2022react} format, where the agentic LLM iteratively generating text and tool-call actions.

\paragraph{Action Space}
The total action space $\mathcal{A}$ decomposes as:
\begin{equation}
\mathcal{A} = {\mathcal{A}_{\text{gen}}} \cup {\mathcal{A}_{\text{tool}}}
\end{equation}
where:
\begin{itemize}
    \item {$\mathcal{A}_{\text{gen}} = \mathcal{V}^*$}: All possible token sequences (LLM-generated text)
    \item {$\mathcal{A}_{\text{tool}} = \{(f, \mathbf{args}) | f \in \mathcal{F}, \mathbf{args} \in \text{Dom}(f)\}$}: Pre-defined set of tool calls
\end{itemize}

\paragraph{Generation Action}
A generation action $a_t^{\text{gen}} \in \mathcal{A}_{\text{gen}}$ is an autoregressively generated token sequence by the LLM:
\begin{equation}
a_t^{\text{gen}} = (w_1, w_2, \ldots, w_n), \quad w_i \in \mathcal{V}
\end{equation}
with generation probability:
\begin{equation}
{\pi_\theta^{\text{gen}}(a_t^{\text{gen}} | \mathcal{H}_t) = \prod_{i=1}^{n} P_\theta(w_i | \mathcal{H}_t, w_{<i})}
\end{equation}

\paragraph{Tool Call Action}
A tool call action $a_t^{\text{tool}} = (f_k, \mathbf{args})$ consists of:
\begin{itemize}
    \item Tool selection: $f_k \in \mathcal{F} = \{f_1, \ldots, f_K\}$
    \item Argument generation: $\mathbf{args} = \{\text{arg}_1, \ldots, \text{arg}_m\}$
\end{itemize}
with generation probability:
\begin{equation}
{\pi_\theta^{\text{tool}}(a_t^{\text{tool}} | \mathcal{H}_t) = P_\theta(f_k | \mathcal{H}_t) \cdot \prod_{j=1}^{m} P_\theta(\text{arg}_j | \mathcal{H}_t, f_k, \text{arg}_{<j})}
\end{equation}

\subsubsection{Environments Building}
To facilitate the agentic training for our targeted tool-use and coding tasks, we develop execution environments that bridges the gap between the LLM's textual outputs and the functional execution of tools. Our built environments provide interfaces to both real code execution engines (Python, JavaScript, \textit{etc}.) and a comprehensive set of tools. Each tool, whether real or simulated, is exposed to the agent as a callable function with a rigorously defined signature, including its name, description, parameter types, and return structure. The simulated tools, which model APIs, system commands, and complex software functions, rely on mock backends implemented as either deterministic rule-based systems or LLM-based agents. For genuine tools such as code execution engines, these backends not only return contextually appropriate outputs based on the input commands but also incorporate realistic error handling (\textit{e.g.}, invalid inputs, resource not found, or \textit{etc}.) to teach robust error recovery. The results, successes, or errors are fed back to the policy model as textual observations, creating a closed-loop learning system where the agent can iteratively refine its actions based on executable feedback, thereby grounding its reasoning in a dynamic and verifiable reality.

\begin{figure}[tb]
    \centering
    \includegraphics[width=\linewidth]{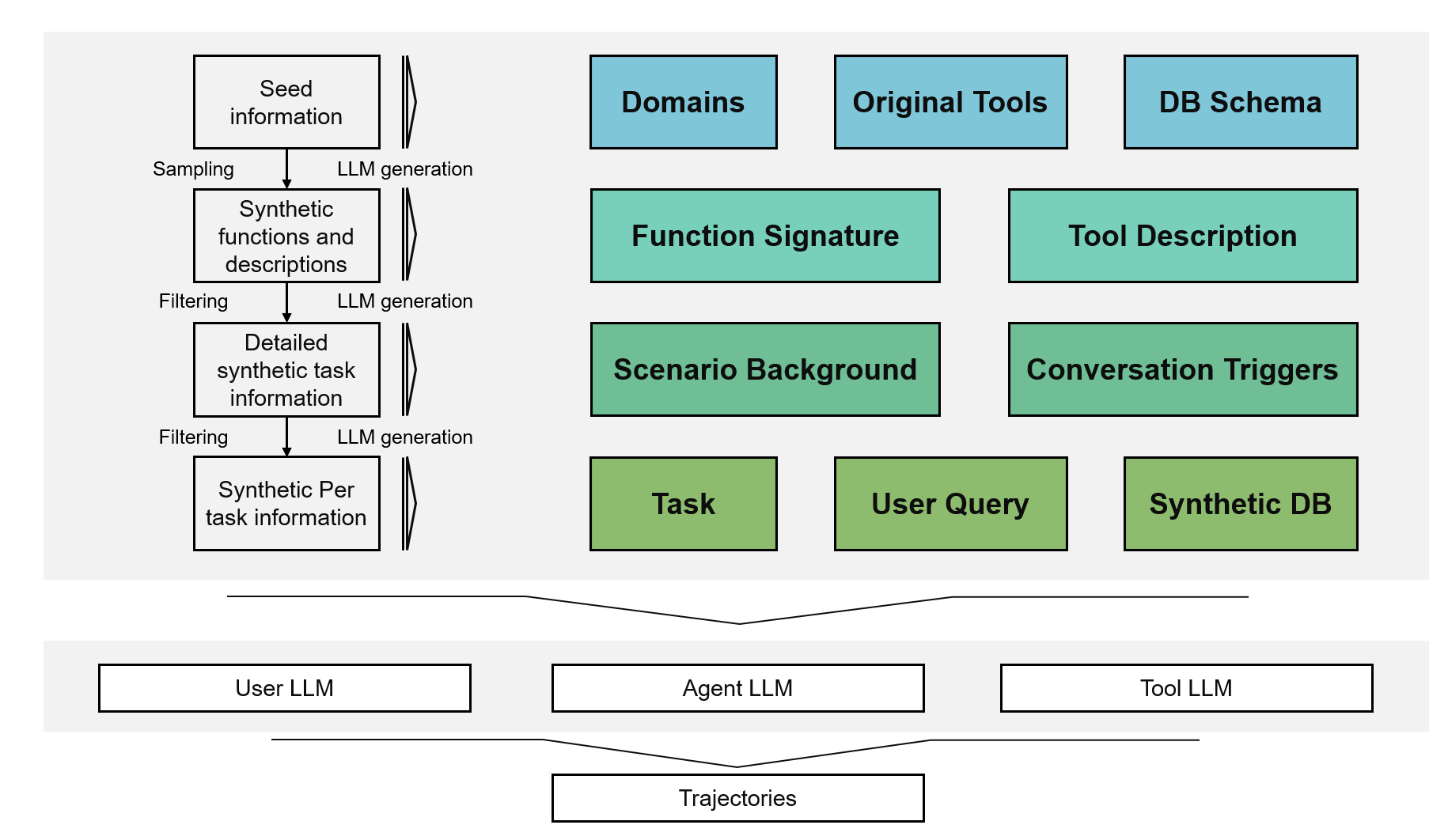}
    \caption{Our multi-turn prompting pipeline for synthesizing tool-use data.}
    \label{fig:tool_syn}
\end{figure}

\subsection{Supervised Fine-Tuning}
\label{subsec:sft}

\subsubsection{Prompt Collection and Trajectory Synthesis}
We start with SFT before the RL training procedure.  
We include around 2.4B tokens, from either open-source high-quality datasets or self-constructed synthetic data, for agentic tool use and coding domains.
\paragraph{Tool-use data} We primarily leverage the tool-calling dataset~\citep{bercovich2025llamanemotronefficientreasoningmodels}. We use various strong LLMs to distill additional trajectories from the same amount of queries.  Additionally, we synthesize more tool-use data using our designed data generation pipeline. Leveraging existing data domains, tools, and database schema, we employ multi-turn prompting with powerful LLMs, illustrated in Figure~\ref{fig:tool_syn}, to generate new tools, tasks, and conversational content. 
For our new distilled and generated trajectories from synthetic conversations, we employ multiple strategies for filtering, including consistency checks within one query, and assessments of each step's format and logical correctness within trajectories.

\paragraph{Coding} Our data incorporates both code contests' problems and software engineering (SWE) tasks from GitHub repositories.
\begin{itemize}
\item 
    For code contests' problems, we build task-solution-test case triplets using data from CodeContests~\citep{li2022competition} and Exercism~\footnote{\url{https://github.com/exercism}}. This process primarily follows the methodology established in the Aider-Polyglot benchmark~\footnote{\url{https://aider.chat/docs/leaderboards/}}. The corresponding trajectories are obtained by first assembling the problem description, response format, few-shot exemplars, and initial project code into a structured input message. We obtain its responses from both the target model (\textit{i.e.}, Qwen3 8B) and a set of stronger LLMs. We verify the generated responses using all test cases, and keep the incorrect responses from Qwen3 8B and correct responses from strong LLMs. We construct the second-turn data by incorporating the failed responses from Qwen3 8B and the compilation/execution error messages into the input messages, then prompting the strong LLMs and retaining those responses that pass verification. In such cases, we obtain one/two-turn level data with the tool calling of corresponding coding language execution engines. Note that the test cases are reserved exclusively for validating responses. To prevent data contamination, the test cases are excluded from the tool inputs during the construction of training trajectories.
\item For GitHub repository data, we make use of the SWE-Smith dataset~\citep{yang2025swe}. The dataset applies a range of controlled perturbation strategies—such as AST-level transformations, patch inversion, templated mutations, and LLM-generated injections—to systematically introduce errors in syntax, semantics, logic, API misuse, and performance regressions, thereby producing paired software-editing data of ``buggy code — fix patch (diff) — tests passing''. We filter for problems that include synthesized problem descriptions, yielding 12,000 problems in total.

For each problem, we distill new trajectories by adopting our developed \textit{mini-swe-agent-plus} as the scaffold.
It is a lightweight adaptation of the mini-swe-agent~\citep{yang2025swe}, which consists of approximately one hundred lines of code and has minimal dependencies, making it easy to debug, extend, and deploy in containerized environments. Compared to more comprehensive agent frameworks like SWE-agent~\citep{yang2024swe} and OpenHands~\citep{wang2024openhands}, there are two engineering features of mini-swe-agent: (i)
\textit{Single Tool and Linear Trajectory}: By default, it only utilizes the bash tool and maintains a linear dialogue history. It does not rely on model-side ``tool calling'' interfaces, thereby reducing dependency on specific LLMs or SDKs;
(2) \textit{Stateless Execution and Easy Sandboxing}: Each action is executed independently via subprocess.run, making it easy to replace with isolated backends such as docker exec or podman. This design supports batch evaluation and reward calculation. 

Based on case analysis, we observe that mini-swe-agent’s reliance solely on bash presents significant limitations in code editing tasks. For example, sed processes text on a line-by-line basis, requiring additional techniques for multi-line replacements, which is both cumbersome and fragile. To address this shortcoming, we introduce a ``string replacement'' editing tool, resulting in \textit{mini-swe-agent-plus}. This enhancement provides atomic and controllable multi-line editing capabilities, improving success rates and reducing ineffective attempts. As shown in Table~\ref{tab:swe_verified_agent}, introducing the editing tool substantially narrows the performance gap between \textit{mini-swe-agent} and \textit{SWE-agent}. On stronger models like \textit{Claude Sonnet~4}, their performance is comparable. The added string-edit operation also noticeably reduces average execution turns, with replacement-style edits used in over 95\% of code-modifying tool calls. 

\begin{table}[t]
    \centering
    \small
    \setlength{\tabcolsep}{0.18in}
    \begin{tabular*}{\textwidth}{@{\extracolsep{\fill}}l|lcc@{}}
\toprule
\textbf{Model} & \textbf{Agent Framework} & \textbf{SWE-bench Verified } & \textbf{Avg. Turns} \\
\midrule
Claude Sonnet 4 & Claude (proprietary) & 72.7\% & - \\
Claude Sonnet 4 & OpenHands             & 70.4\% & - \\
Claude Sonnet 4 & SWE-agent             & 66.6\% & - \\
Claude Sonnet 4 & mini-swe-agent        & 64.9\% & 79.1 \\
Claude Sonnet 4 & mini-swe-agent-plus   & 67.0\%  & 67.3 \\
\hline
Claude Sonnet 4.5 & mini-swe-agent      & 70.6\% & - \\
Claude Sonnet 4.5 & mini-swe-agent-plus & 71.8\% & - \\
\hline
Claude Sonnet 3.7 & SWE-agent           & 62.4\% & - \\
Claude Sonnet 3.7 & mini-swe-agent      & 52.8\% & 62.9 \\
Claude Sonnet 3.7 & mini-swe-agent-plus & 54.6\% & 48.7 \\
\bottomrule
    \end{tabular*}
    \caption{SWE-Verified accuracy and average execution turns across models and agent frameworks}
    \label{tab:swe_verified_agent}
\end{table}

During trajectory filtering, we remove those that could not produce a submit-ready patch. We also consider filtering by executing the test environment. However, at our current scale, excluding test-failing cases had a negligible impact on results. To prevent test-set leakage, we filter the training repositories and dropped any that overlap with the test set. At last, the SFT dataset for SWE contains about 66,000 examples. 
\end{itemize}

\subsubsection{SFT Scaling}
We examine how scaling the compute budget for SFT affects performance. We analyze three specific scaling axes: (1) increasing model backbone size, (2) expanding the volume of agentic SFT data, and (3) augmenting the data with a combination of agentic and other reasoning tasks.

To explore these scaling strategies, we construct three pilot SFT experiments. First, we finetune from 8B and 32B models to compare the model scaling.
Next, we construct various datasets to examine the effect of scaling the data size.
At last, we increase the amount of SFT data by incorporating the use of long chain-of-thought reasoning data.
In Section~\ref{subsec:discussion_sft}, we provide our SFT results with the purpose to answer the following questions:
\begin{enumerate}
    \item How does scaling the compute by enlarging the model parameters improve model performance?
    \item How does scaling agentic SFT data improve model performance on agentic tasks? What is the most important factor to scale up the data?
    \item Is scaling the compute in the reasoning abilities helpful for agentic abilities?
\end{enumerate}

\subsection{Reinforcement Learning}
\label{subsec:rl}

\subsubsection{Training Objective}
\label{subsec:overview}
We generate $G$ rollouts $\{o_i\}^{G}_{i=1}$ for each question $q$ within a global batch of $N$ prompts, then execute a single policy gradient update using $M$ prompts, where $M < N$. Given an LLM $\pi_{\theta}$ parameterized by $\theta$, we apply the revised GRPO~\citep{shao2024deepseekmath} objective formulated as:

\begin{equation}
\begin{aligned}
\mathcal{J}(\theta) 
= \; &\mathbb{E}_{(q,a)\sim\mathcal{D},\, \{o_i\}_{i=1}^G\sim \pi_{\text{inference}}(\cdot\mid q)} \Bigg[
\frac{1}{\sum_{i=1}^{G}|o_i|}\sum_{i=1}^{G}\sum_{t=1}^{|o_i|}
\min\!\left(
\frac{\pi_{\text{training}}(o_{i,t}\!\mid\! q, o_{i,<t},\theta_{\text{old}})}{\pi_{\text{inference}}(o_{i,t}\!\mid\! q, o_{i,<t},\theta_{\text{old}})},\, C
\right)
\\
&\qquad\qquad\cdot
\min\Big(
r_{i,t}(\theta)\hat{A}_{i,t},\;
\text{clip}\!\big(r_{i,t}(\theta),\, 1-\varepsilon_{\text{low}},\, 1+\varepsilon_{\text{high}}\big)\hat{A}_{i,t}
\Big)
\Bigg],
\label{eq:grpo_tis_obj}
\end{aligned}
\end{equation}
where a question-answer pair $(q, a)$ is sampled from the training dataset $\mathcal{D}$, $\{o_i\}^{G}_{i=1}$ are responses generated for $q$ by the current policy $\pi_\theta(\cdot | q)$, $\varepsilon_\text{low}$ and $\varepsilon_\text{high}$ denote the lower and upper clipping ratios.
$r_{i,t}$ and $\hat{A}_{i,t}$ are formulated respectively as follows:
\begin{equation}
    r_{i,t}(\theta)=\frac{\pi_{\theta}(o_{i,t} \mid q, o_{i,<t})}{\pi_{\theta_{\text{old}}}(o_{i,t} \mid q,o_{i,<t})},\quad\hat{A}_{i,t} = \frac{R_i - \text{mean}(\{R_i\}_{i=1}^G)}{\text{std}(\{R_i\}_{i=1}^G)}.
\label{eq:advantage_calculation}
\end{equation}
Our approach applies the following key modifications to the original GRPO objective:
\begin{itemize}
\item Removal of the KL regularization term;
\item Asymmetric DAPO clipping~\citep{yu2025dapo};
\item Truncated importance sampling to correct for distributional discrepancy between the inference policy $\pi_{\text{inference}}$ and the training policy $\pi_{\text{training}}$~\citep{yaoyour};
\item Oversampling and masking of trajectories that reach the maximum context, step count, or timeout~\citep{deepswe2025}.
\end{itemize}

\subsubsection{Mixing the Use of Turn-Level and Outcome Rewards}
We mix the use of two RL strategies during training. The first is end-to-end multi-turn RL. In this approach, the model generates an entire action trajectory before receiving a reward based on the final outcome. This allows the model to perform iterative optimization of its policy through interactions with the tools, thereby significantly enhancing its decision-making abilities. In this setting, $R_i$ refers to the reward for the whole trajectory rollout $o_i$, which is computed as:
\begin{equation}
R_i =
\left\{
\begin{array}{ll}
1, & \text{if } \text{FormatCorrect}(o_i) \text{ and } \text{TaskCompleted}(q, o_i), \\[6pt]
0, & \text{otherwise.}
\end{array}
\right.
\label{eq:outcome_calculation}
\end{equation}
where $\text{TaskCompleted}(q, o_i)$ is a binary indicator of task success, determined by the environment's feedback. This feedback can be based on a comparison of the database state against a ground truth or the pass rate of test cases executed by an engine.

However, end-to-end multi-turn RL is often hampered by reward sparsity. That is, very few generated trajectories can be assigned with positive outcome rewards. To mitigate this, we introduce a second strategy: step-wise RL. We identified a subset of prompts with trajectories where the correct sequence of tool calls is deterministic and known. For these prompts, we train the model to generate the next assistant response (be it a function call or a user-facing message) given the task context and previous tool calls. By applying rule-based, turn-level rewards, we provide more frequent guidance, steering the model toward making correct function calls at each step. A reward of 1 is awarded only if the $T$-th tool call is both correctly formatted and its content matches the ground truth exactly, including the tool name, all parameters, and every data field:
\begin{equation}
R_{i,T} =
\left\{
\begin{array}{ll}
1, & \text{if } \text{FormatCorrect}(o_{i,T}) \text{ and } \text{ExactMatch}(o_{i,T}, o_{i,T}^*), \\[6pt]
0, & \text{otherwise.}
\end{array}
\right.
\label{eq:turn_calculation}
\end{equation}
where $o_{i,T}^*$ is the given ground truth tool call.

\subsubsection{Training Implementation}
Agentic tasks, characterized by long  trajectories with frequent environmental interaction, introduce significant training efficiency challenges compared to single-turn generation. 
A critical issue in agentic RL training is GPU idling, where resources for short trajectories remain underutilized while waiting for long ones to finish~\citep{gao2025rollpacker,zhang2025agentrlscalingagenticreinforcement}. This imbalance severely degrades training efficiency and scalability, necessitating an asynchronous training framework.
Another issue is the concurrent deployment of many environments to support rollouts for collecting multi-turn feedback for different tasks.
Therefore, we utilize the following training implementation to improve efficiency in multi-turn RL training:
\begin{itemize}
  \item \textit{Disaggregated architecture.} Instead of utilizing the popular colocated architecture, we disaggregate compute into $N_r$ rollout nodes and $N_\ell$ train nodes. Train nodes continuously consume batches while rollout nodes keep generating trajectories, overlapping training with ongoing rollouts.
  \item \textit{Asynchronous reward.} Instead of waiting for all rollouts to finish before rewarding, the final interaction with the sandbox returns a reward payload (\textit{e.g.}, pass/fail score) immediately. Rollout workers enqueue \((s_t, a_t, r_t,\dots)\) per-sample, eliminating batch-level barriers and reducing straggler-induced bubbles.
  \item \textit{Per-task sandbox.} Each task (\textit{e.g.}, SWE, tool use) runs its tool environment in its own containerized sandbox. Each sandbox exposes an RPC and tool requests will be dispatched to the right sandbox accordingly. It guarantees the disentanglement between training and tool execution, leaving room for scaling the sandbox and train resources independently.
\end{itemize}

\subsubsection{RL Experiment Roadmap}
Our goal is to develop a unified agentic LLM capable of handling diverse tool calls and environment interactions across multiple tasks and domains. A natural approach is to perform joint RL training on these tasks. However, multi-task agentic RL presents several practical challenges: domains and tasks differ in their state/action spaces, environment execution times, as well as task complexity and rollout length budgets. Consequently, naive joint training can lead to a scenario where one task shows significant improvement while another stagnates, resulting in training instability and performance imbalance~\citep{wang2025uitars2technicalreportadvancing,zhang2025agentrlscalingagenticreinforcement}. Therefore, we adopt a simpler but effective strategy: starting from a shared SFT initialization, we conduct separate RL training tailored to different environments. For example, our chosen tasks (tool-use, coding contests, and SWE) involve almost non-overlapping environment interactions. We then merge these trained models using Select–Calculate–Erase~\citep{wan2024fusechatknowledgefusionchat}, which forms task vectors relative to a shared base model and performs variance-based selection, matrix-level energy weighting, and sign-consensus filtering before adding the normalized aggregate update back to the base:
\begin{align}
\Delta_i &= \theta_i - \theta_{\text{base}}, \quad i=1,\dots,K, \\
w_i &= \frac{\|\Delta_i\|_F^2}{\sum_{j=1}^{K}\|\Delta_j\|_F^2},
\qquad \sum_{i=1}^{K} w_i = 1,\; w_i \ge 0, \label{eq:weights}\\[4pt]
\theta_{\text{fused}}
&= \theta_{\text{base}}
  + \Big(\sum_{i=1}^{K} w_i\, \Delta_i\Big)
    \odot C \odot M. \label{eq:sce}
\end{align}

\noindent where $\theta_i$ are the $K$ model parameters, $\theta_{\text{base}}$ is the base model,
$\Delta_i=\theta_i-\theta_{\text{base}}$ are task vectors,
$w_i=\|\Delta_i\|_F^2\big/\sum_{j=1}^K\|\Delta_j\|_F^2$ are normalized energy weights,
$C\in\{0,1\}^d$ is the sign-consensus mask,
$M\in\{0,1\}^d$ is the variance Top-$p$ mask.

Beyond RL which improves the model’s internal decision-making at inference time, we further explore test-time scaling through parallel generation~\citep{snell2024scalingllmtesttimecompute}. In this setup, the agentic LLM generates multiple candidate trajectories for the same input prompt, and a verifier or selection strategy determines the final output. This formulation enables us to analyze how increasing test-time computation and employing different verification strategies jointly influence the overall performance of agentic LLMs.
As discussed in~\cite{chen2024more}, the empirical gains from test-time scaling often fall short of the theoretical upper bound, largely due to the difficulty of designing effective verifiers that can accurately identify the correct trajectory among multiple candidates. To better understand this gap, our work provides a systematical analysis of various verification strategies, assessing their adaptability and potential on agentic tasks.

Based on the preceding discussion, the following questions regarding the RL process are explored in Section~\ref{subsec:discussion_rl}:
\begin{enumerate}
    \item What does the training dynamics reveal about the stability of current agentic RL methods?
     \item What is the impact of the disaggregated framework on training efficiency?
     \item When building a unified agentic LLM for multiple tasks, which strategy is more effective: multi-task joint RL training or a combination of single-task RL training followed by model merging?
     \item Can test time scaling strategies offer further performance gains?
\end{enumerate}

\section{Experiments}
\label{sec:experiments}

\subsection{Experimental Setup}
\label{subsec:evaluation}
The model is evaluated on the following  tool-use and coding tasks: 
\begin{itemize}
    \item BFCL v3~\citep{patil2025bfcl}: the Berkeley Function-Calling Leaderboard extends BFCL-v1 (expert-curated, single-turn) and BFCL-v2 (user-contributed/live, single-turn) by introducing a \emph{multi-turn, multi-step} category. Beyond choosing a single function, models must plan across turns using realistic API suites (\textit{e.g.}, file system, travel booking, trading, vehicle control), with success judged by the \emph{post-execution system state} rather than parameter matching. BFCL v3 provides a broad measure of tool-use reasoning and robustness. Our inference configuration is set with temperature = 0.7, top\_p = 0.95, max\_length = 16k.
    \item $\tau$-bench~\citep{yao2024tau}: a benchmark for tool-agent-user interaction, which includes evaluation scenarios in two real-world domains: Retail and Airline. In each domain, $\tau$-bench uses a large language model (typically GPT-4o) to simulate the user, while the model under evaluation serves as the service agent. The agent is provided with a set of tools that enable it to fulfill the user's requests. It should invoke the appropriate tools based on user's needs to operate on a provided database. During the evaluation, the agent is considered to have completed a task correctly when, after its operations, the database exactly matches the ground truth. We use $\text{temperature}=0.8$ and keep the other sampling parameters at their default values. Since $\tau$-bench exhibits substantial run-to-run variability, we report avg@4 results for our models to mitigate variance.
    \item SWE-bench Verified~\citep{jimenez2024swebench}: a human-validated subset of SWE-bench containing 500 real GitHub issues that assess whether systems can produce patches resolving bugs and passing all tests. For evaluation, we use the mini-swe-agent-plus scaffold, capping each run at 200 steps with a 64k context length.
    \item Aider Polyglot: a benchmark to evaluate an LLM’s ability to follow instructions and edit code successfully without human intervention, which tests LLMs on 225 challenging Exercism coding exercises across C++, Go, Java, JavaScript, Python, and Rust. Our inference configuration is set with temperature = 0.7, top\_p = 0.8 and max\_length = 24K.

\end{itemize}

\subsection{Main Results}
\label{subsec:main_results}

\begin{table}[t]
    \centering
    \small
    \setlength{\tabcolsep}{0.2in}
    \begin{tabular*}{\textwidth}{@{\extracolsep{\fill}}l|ccc@{}}
\toprule
     \textbf{Model} &  \textbf{BFCL v3} & \textbf{$\tau$-bench Retail} & \textbf{$\tau$-bench Airline}  \\
     \midrule
     Claude Sonnet 4.0 [\citenum{claudesonnet-4}] & 73.3 & 80.5 & 60.0 \\
     OpenAI GPT-4.1 [\citenum{gpt4.1}] & 62.1 & 68.0 & 49.4 \\
     Qwen3-Coder-480A35B-Instruct [\citenum{qwen3technicalreport}] & 68.7 & 77.5 & 60.0 \\
     GLM 4.5 [\citenum{zeng2025glm}] & 77.8 & 79.7 & 60.4 \\
     Kimi K2 [\citenum{team2025kimi}] & 71.1 & 73.9 & 51.2 \\ \hline
     Qwen3 Next-80A3B-Instruct & 70.3 & 60.9 & 44.0 \\
     Qwen3-30A3B-2507-Instruct & 65.1 & 59.1 & 40.0 \\
     Qwen3-30A3B-2507-Thinking & 72.4 & 67.8 & 48.0 \\
     Qwen3-32B (Thinking) & 70.4* & 48.7* & 28.0* \\
     Qwen3-32B (Non-Thinking) & 62.8* & 40.0* & 20.0* \\ \hline
     Qwen3-8B (Thinking) & 68.2* & 45.2 & 25.0 \\
     Qwen3-8B (Non-Thinking) & 59.8* & 35.7* & 12.0* \\
     xLAM-2-8B-fc-r [\citenum{prabhakar2025apigen}] & 72.8 & 58.2 & 35.2 \\
     AgentScaler-8B [\citenum{fang2025towards}] & - & 50.4 & 42.0 \\ \hline
     \bf{Klear-AgentForge-8B} & 71.5 & 56.7 & 41.5 \\
\bottomrule
    \end{tabular*}
    \caption{Agentic Tool use results. Results marked with $*$ indicate that no official performance was available in the reference. For these, we applied our own evaluation script to ensure a consistent comparison.}
    \label{tab:result_tool_use}
\end{table}

\begin{table}[htbp]
    \centering
    \small
    \setlength{\tabcolsep}{0.2in}
    \begin{tabular*}{\textwidth}{@{\extracolsep{\fill}}l|cc@{}}
\toprule
     \textbf{Model} &  \textbf{SWE-bench verified} & \textbf{Aider-Polyglot} \\
     \midrule
     Claude Sonnet 4.0 & 72.7 & 56.4\\
     Qwen3-Coder-480A35B-Instruct & 69.6 & 61.8 \\
     Deepseek-v3.2-exp & 67.8 & 74.5 \\
     Kimi-K2 & 65.8& 60.0 \\
     OpenAI GPT-4.1 & 63.8 & 52.4 \\ \hline
     Qwen3-32B (Thinking) & 23.0 & 40.0 \\
     Qwen3-32B (Non-Thinking) & 18.8* & 41.3 \\
     Qwen3 Next-80A3B-Instruct & 16.0*& 49.8 \\
     Qwen3-30A3B-2507-Instruct &17.6* & 35.6 \\
     Qwen3-30A3B-2507-Thinking & 10.8*& 15.1* \\ 
     Qwen3-Coder-30A3B-Instruct & 51.6 & 33.3 \\
     \hline
     DeepSWE-32B-Preview [\citenum{deepswe2025}] & 42.2 & - \\
     Skywork-SWE-32B [\citenum{skywork-swe}]  & 38.0 & - \\
     SWE-Swiss-32B-SFT [\citenum{SWESwiss2025}]  & 36.0 & - \\
     SWE-Swiss-32B-RL  & 45.0 & - \\
     \hline
     Qwen3-8B (Thinking) & 9.8*& 17.3* \\
     Qwen3-8B (Non-Thinking) &8.0* & 16.4* \\
     SWE-Mirror-LM-7B [\citenum{wang2025swemirrorscalingissueresolvingdatasets}] & 22.8 & - \\
     SWE-agent-LM-7B [\citenum{yang2025swesmith}] &15.2 & - \\ \hline
     \bf{Klear-AgentForge-8B} &  39.4 & 33.8 \\
\bottomrule
    \end{tabular*}
    
    \caption{Agentic coding results.}
    \label{tab:result_coding}
\end{table}

We first present the main benchmark results for our model, Klear-AgentForge-8B. It is derived from the Qwen3-8B base model and trained using a pipeline of SFT, RL, and final model merging. Results on tool use and coding benchmarks are shown in Table~\ref{tab:result_tool_use} and Table~\ref{tab:result_coding} respectively. Our Klear-AgentForge-8B model significantly outperforms the official posttrained Qwen3-8B models in both Thinking and Non-Thinking modes. It also demonstrates highly competitive performance against other agentic LLMs derived from Qwen3/Qwen2.5. Crucially, our model is uniquely versatile, excelling across multiple agentic tasks, while the compared agentic models (e.g. AgentScaler-8B, SWE-Mirror-LM-7B) are largely confined to only one domain such as tool use or coding.
For instance, on the BFCL v3 benchmark, Klear-AgentForge-8B demonstrates consistent, high performance across Live, Non-Live, and Multi-turn categories, showcasing its balanced proficiency in both synthetic and user-driven function calls.
On coding benchmarks such as SWE-bench Verified, Klear-AgentForge-8B significantly outperforms other 8B models and even matching the performance of many open 32B systems (e.g., Skywork-SWE-32B and SWE-Swiss-32B-SFT). This demonstrates the effectiveness of our training data construction and the designed training recipe.

\subsection{SFT Analysis}
\label{subsec:discussion_sft}

\subsubsection{Model Scaling}
We first investigate how model scale influences agentic capabilities by evaluating models
with 8B and 32B parameters on $\tau$-bench and SWE-Bench (verified).
We evaluate performance relative to the released 8B and 32B non-thinking counterparts. Results are shown in Table~\ref{tab:result_modelscaling}. Unsurprisingly, the 32B SFT model achieve higher overall performance than its 8B counterpart. However, we observe a more significant performance gain on the 8B models. This indicates that small language model backbones can rapidly enhance their agentic capabilities through in-domain data finetuning, underscoring the value of further research into building small models for agentic tasks, as also suggested by \cite{bercovich2025llamanemotronefficientreasoningmodels}.

\begin{table}[t]
    \centering
    \small
    \setlength{\tabcolsep}{0.2in}
    \begin{tabular*}{\textwidth}{@{\extracolsep{\fill}}l|ccc@{}}
\toprule
     \textbf{Model} &   \textbf{$\tau$-bench Retail} & \textbf{$\tau$-bench Airline} & \textbf{SWE Bench Verified} \\
     \midrule
         Qwen3-8B (Non-Thinking) & 35.7* & 12.0* & 8.0*\\
         Klear-AgentForge-8B-SFT & 54.8 & 32.0 & 38.2 \\
     Qwen3-32B (Non-Thinking) & 40.0* & 20.0* & 23.0* \\
     Klear-AgentForge-32B-SFT & 59.8 & 37.5 & 55.2 \\ 
\bottomrule
    \end{tabular*}
    \caption{SFT results on model scaling.}
    \label{tab:result_modelscaling}
\end{table}

\subsubsection{Data Scaling}

\begin{figure}[t]
    \centering
    \includegraphics[width=0.85\linewidth]{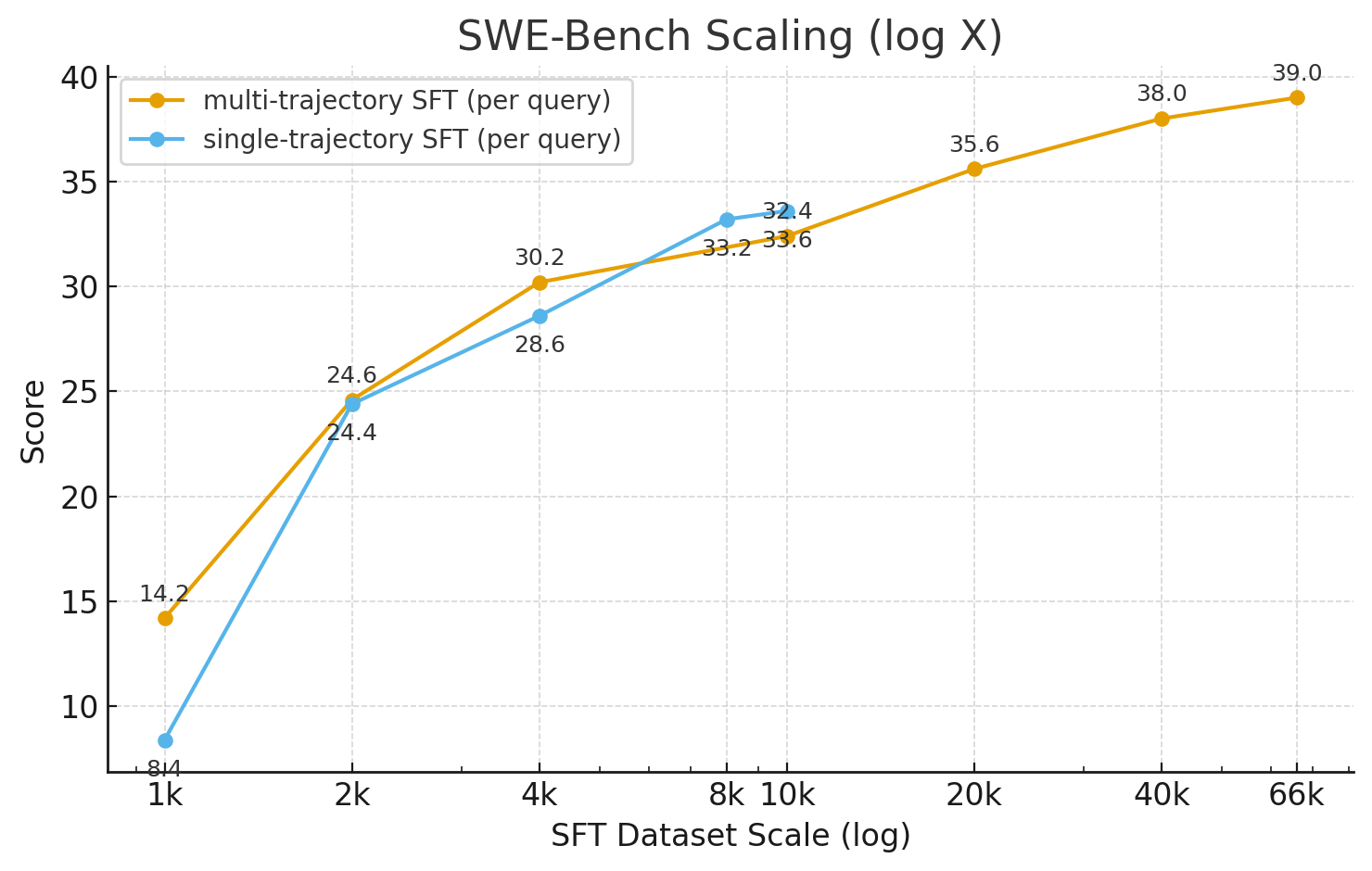}
    \caption{Comparison of scaling effects between single-trajectory and multi-trajectory SFT data on SWE-Bench (verified).}
    \label{fig:sft_scaling_accuracy}
\end{figure}

We conduct a pivot study using the SWE SFT dataset and evaluate performance on the SWE-Bench (verified) benchmark.
To examine the impact of scaling, we consider two settings:  
(1) Single-trajectory per query, where each prompt has exactly one trajectory; and  
(2) Multi-trajectory per query, where each prompt may have multiple trajectories, and samples are randomly drawn from the full set.
As shown in Figure~\ref{fig:sft_scaling_accuracy}, the two settings achieve comparable performance across different dataset scales. This suggests that, at this stage, both increasing the number of unique prompts and increasing the number of trajectories per prompt can lead to similar gains in model accuracy.  
Given this observation, we adopt the simpler multi-trajectory setting for further scaling, which continues to yield steady improvements as the dataset grows.

\subsubsection{Incorporating Reasoning SFT data}
To study the link between reasoning and agentic capabilities, we apply SFT to a reasoning model  rather than a base LLM.
To save our computation efforts, we use Deepseek-distill-qwen3-8B~\citep{deepseekai2025deepseekr1incentivizingreasoningcapability} as our initial model and continue to train it with the same SFT data (48k context length). Unfortunately, we observe a significant performance drop across all evaluation benchmarks. Almost all decrease to 0. Manual analysis revealed the primary cause: the model generates excessively long thinking trajectories before a tool call action, consistently exhausting the 64k output limit without achieving a task-complete state. This result suggests that reasoning and agentic capabilities may not directly enhance each other. Therefore, we must carefully design the data mix and training strategies to develop a model that excels in both.

\subsection{RL Analysis}
\label{subsec:discussion_rl}

\subsubsection{Training Dynamics}
\begin{figure}[t]
    \centering
    \begin{minipage}[t]{0.38\columnwidth}
        \centering
        \includegraphics[width=\textwidth,height=0.165\textheight]{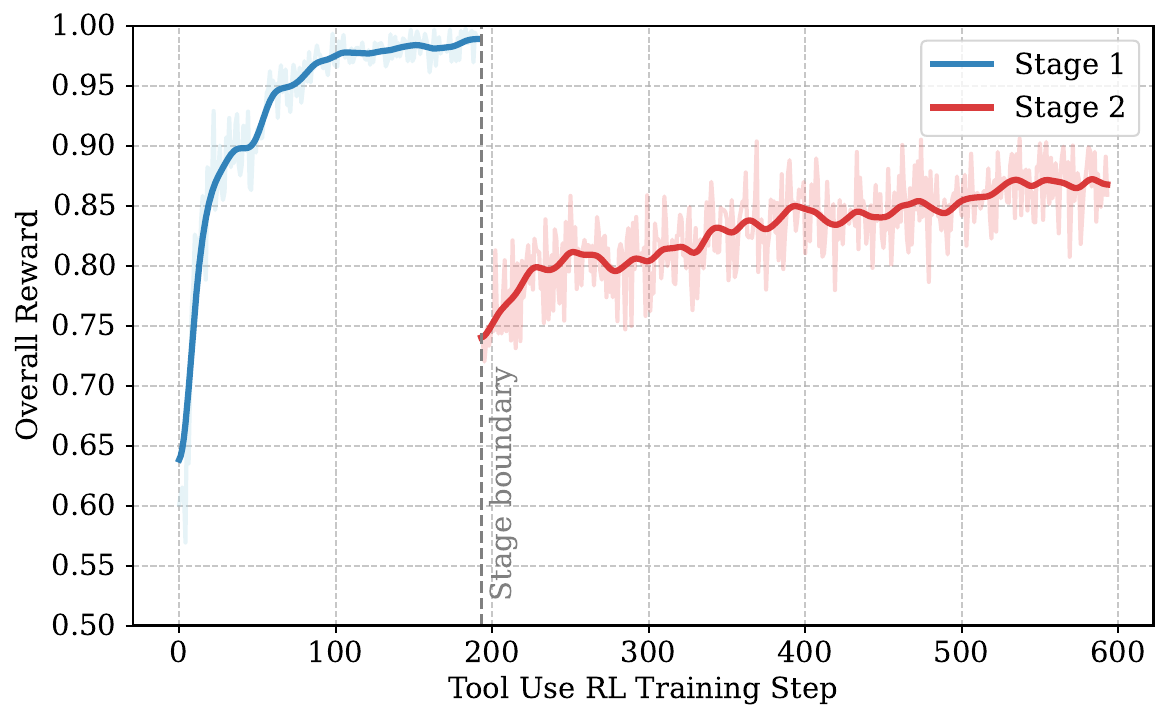}
    \end{minipage}
    \hfill
    \begin{minipage}[t]{0.3\columnwidth}
        \centering
        \includegraphics[width=\textwidth]{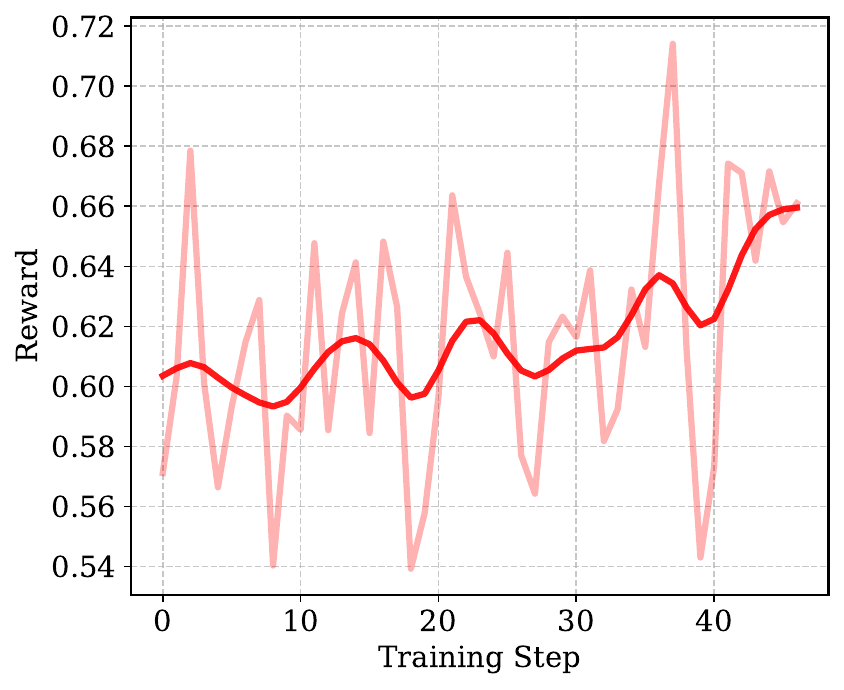}
    \end{minipage}
    \begin{minipage}[t]{0.3\columnwidth}
        \centering
        \includegraphics[width=\textwidth]{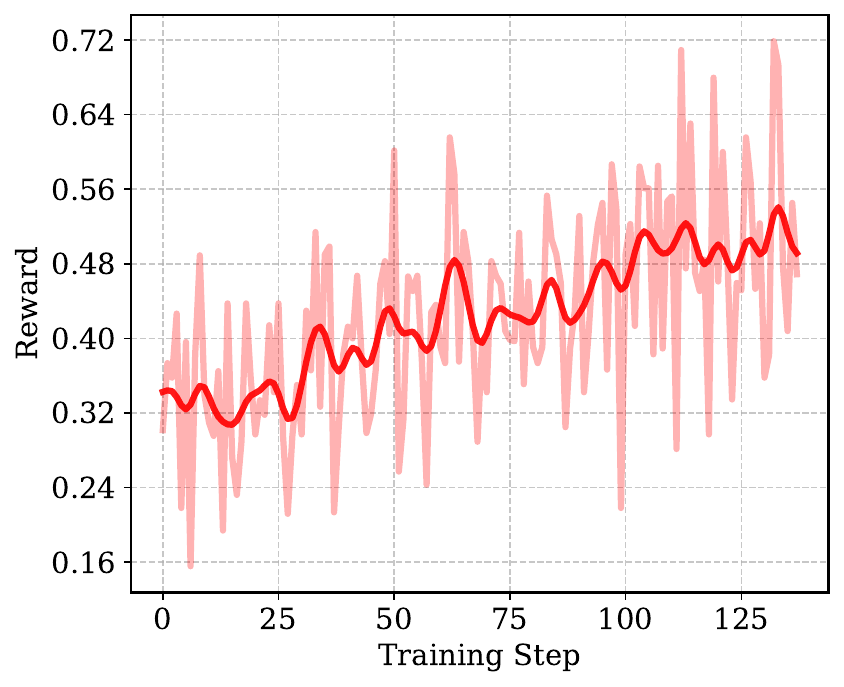}
    \end{minipage}
    \caption{Tool use (Left), SWE (Middle), Code contest (Right) task training reward scores.}
    \label{fig:reward_score}
\end{figure}

We conduct RL on  using the Klear-AgentForge-8B-SFT model. 
For the tool-use domain, the RL process is divided into two stages, as illustrated in Figure~\ref{fig:reward_score} (Left).
In Stage~1, we train the model on the Open-Agentic-Tool-Use dataset\footnote{\url{https://www.modelscope.cn/models/hbg400/Open-Agentic-tool-use}}. To ensure data reliability, we apply a filtering procedure in which each sample is re-evaluated by the Qwen3-235B-A22B-Thinking-250 model, and only those responses that pass the secondary verification are retained for RL training.  
In Stage~2, we continue training on a curated subset of our SFT tool-use dataset. Specifically, the data are filtered based on the success rates measured by the SFT model, focusing on instances that the model has not yet fully mastered (\textit{i.e.}, those with success rates $0 < p < 1$).
As shown in Figure~\ref{fig:reward_score} (Left), the overall reward rises rapidly during Stage~1 and continues to improve steadily throughout Stage~2. This demonstrates that the two-stage RL framework effectively enhances the model’s performance in tool-use reasoning and execution tasks. %

For the SWE domain, 
 SFT samples are filtered with a strong teacher model, and we retain only those that can be solved at least once, ensuring that training instances are truly resolvable and that the validation environment is correctly configured. The learning curve is shown in Figure~\ref{fig:reward_score} (Middle). Reward increases slightly during the training and then exhibits a jump after roughly one epoch (around 40 steps), at which point validation performance begins to decline. 

For code contest task, we verify the sampled 8 responses for each prompt and collect the pass rate. Then, we keep only the prompts with  responses with success rates $0.25\leq p \leq 0.75$, as the final training prompt. As shown in Figure~\ref{fig:reward_score}, the training reward continues to improve, while the entropy remains stable without collapsing.

\subsubsection{Disaggregated RL with Steptime Breakdown}
\begin{figure}[t]
  \centering
  \includegraphics[width=\linewidth]{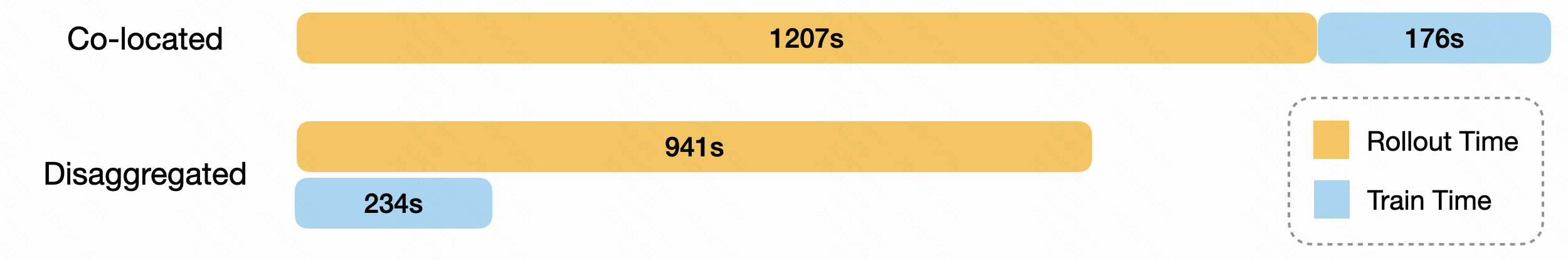}
  \caption{Steptime breakdown on SWE-bench Verified. For disaggregated framework, 4 nodes is used for training and the other 4 nodes for inference.}
  \label{fig:step-time}
\end{figure}

We conduct our RL training on the SWE task using 8 nodes, 64 H800 GPUs in total, for the purpose of efficiency analysis.
Figure~\ref{fig:step-time} shows the per-step breakdown across stages, comparing disaggregated and co-located training framework. By overlapping training phase with rollout phase, disaggregated training framework achieves around 32\% efficiency boost. Note that, compared to 8 nodes co-located, while the training time is longer, the inference time is even shorter. We suspect that with a fixed batch size, there can be some inference resource waste when directly scaling up the number of nodes, especially for long-tail responses which can cause other GPUs idling until those trajectories completed.

\subsubsection{Multi-Task RL vs Model Merge}
\begin{figure}[t]
    \centering
    \includegraphics[width=0.85\linewidth]{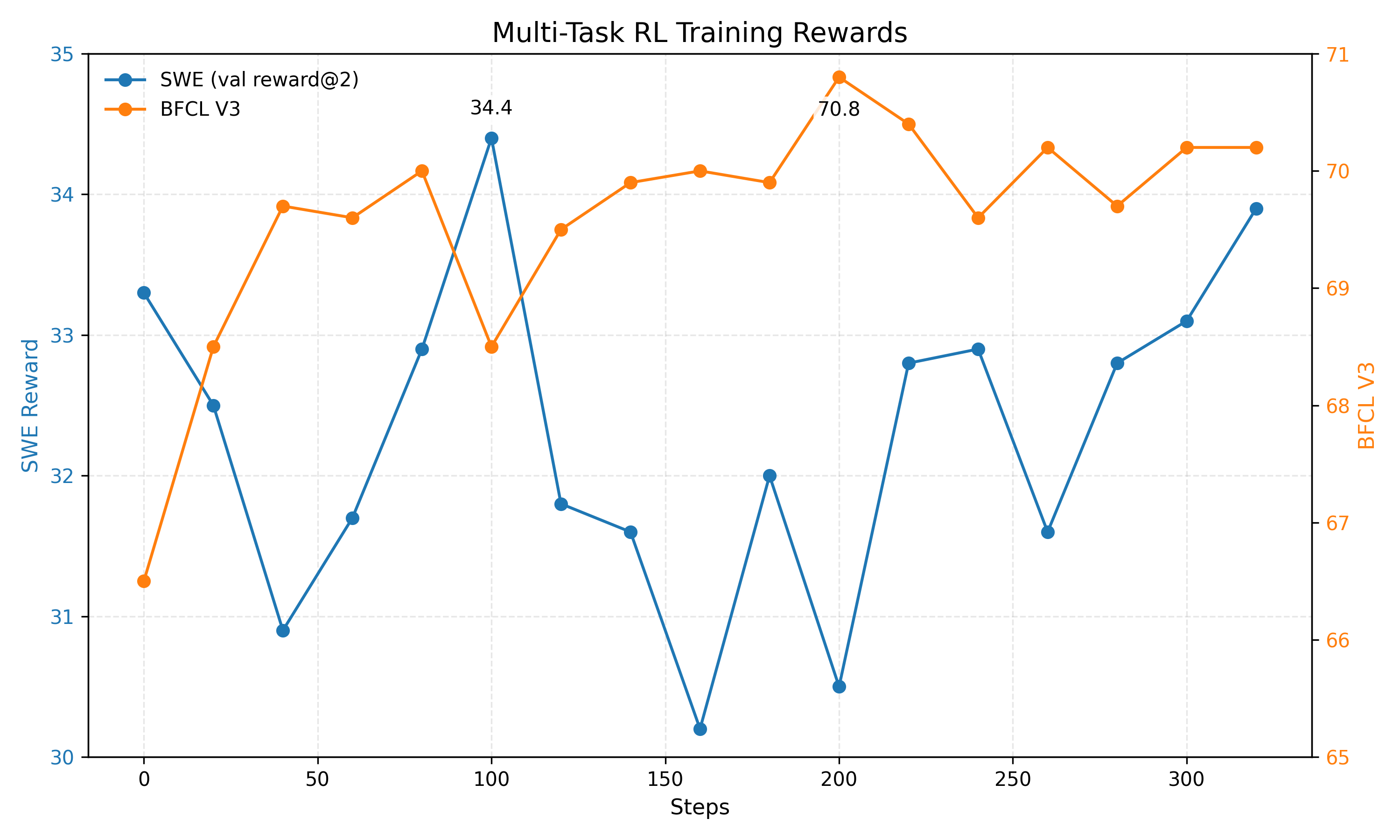}
    \caption{Multi-Task RL training reward scores. (Tool-use uses the v3 benchmark score.)}
    \label{fig:swe_bfcl_v3}
\end{figure}

While model merge serves as our primary approach for consolidating specialized
agents, we also investigate an alternative based on multi-task reinforcement learning. For Multi-Task RL training, we select the tool use and SWE task for experiments. Rather than mixing data randomly, we apply a step-level schedule. This involves a cycle of four steps: three for tool use followed by one for SWE. This schedule maintains the overall 3:1 data ratio found in the full training set, a proportion dictated by the use of Stage 1 tool use data and the entirety of the SWE data. Figure~\ref{fig:swe_bfcl_v3} shows the training rewards for both tasks fluctuating as the policy learns from their interleaved samples. We should say that, the training should be carefully monitored in order to avoid the model training collapse before finishing the update of all our prepared samples. Ultimately, we observe that joint RL training also produces significant performance gains, with BFCL v3 70.8 and SWE-bench verified 39.6. The current results indicate that while this simple joint RL training can nearly match the gains of separate training, it does not produce a synergistic performance boost where the tasks mutually enhance each other.

As for model merge, 
we find that for both coding tasks, the performance drop after model merge (SWE-bench verified: 40.4->39.4, Aider-polyglot: 35.1->33.8). One possible explanation is that we train significantly more samples for tool-use than for coding.
Therefore, despite the efficiency of model merging, multi-task RL training remains essential for the systematic design of data and training strategies tailored to our specific objectives.

\subsubsection{Test-time Scaling}
We evaluate several representative selection strategies, including:
\begin{itemize}
    \item Majority Voting~\citep{brown2024large}: aggregates candidate outputs based on consensus among them;
    
    \item LLM-based selection~\citep{chen2024simple}: leverages both the current model and an auxiliary evaluator (e.g., Qwen32) to select the final output through a knockout tournament involving groupwise candidate comparisons;
    
    \item LogProb Selection~\citep{kang2025scalable}: ranks candidates by their average token-level log-probabilities. For a generated sequence $y = (y_1, \dots, y_n)$ conditioned on input $x$, the average log-probability is
    \[
    \text{AvgLogP}(y|x) = \frac{1}{n} \sum_{i=1}^{n} \log p(y_i \mid x, y_{<i}),
    \]
    where $p(y_i \mid x, y_{<i})$ denotes the model-assigned probability of token $y_i$. A higher $\text{AvgLogP}$ implies greater overall model confidence in the generation;
    
    \item Confidence Selection~\citep{fu2025deep}: measures trajectory quality using metrics derived from internal token distributions. The \textit{token confidence} at position $i$ is defined as
    \[
    C_i = -\frac{1}{k} \sum_{j=1}^{k} \log P_i(j),
    \]
    where $P_i(j)$ is the probability of the $j$-th most likely token, and $k$ (typically 100) is the number of top tokens considered. The \textit{average trace confidence} over all generated tokens is
    \[
    C_{\text{avg}} = \frac{1}{N} \sum_{i=1}^{N} C_i,
    \]
    where $N$ is the total token count. The candidate with the highest $C_{\text{avg}}$ is selected;
    
\end{itemize}

\begin{figure}[h]
    \centering
    \begin{minipage}{0.56\textwidth}
        \centering
        \includegraphics[width=\linewidth]{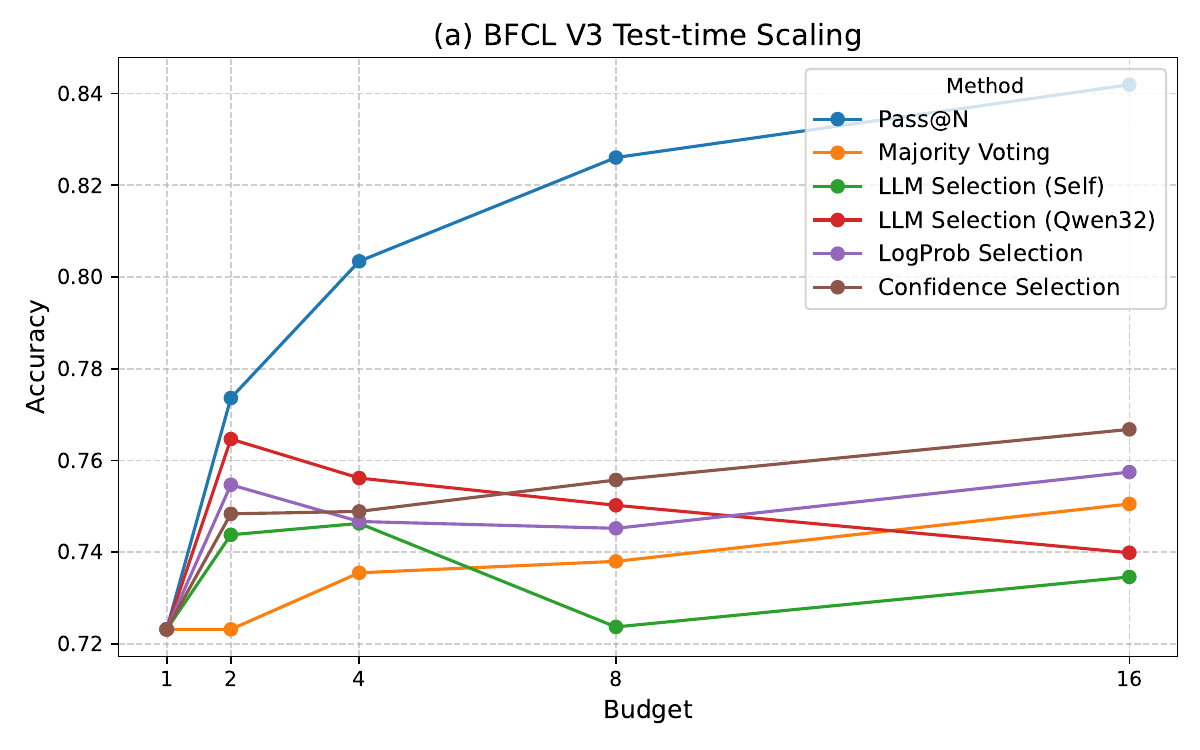}
    \end{minipage}
    \hfill
    \begin{minipage}{0.42\textwidth}
        \centering
        \includegraphics[width=\linewidth]{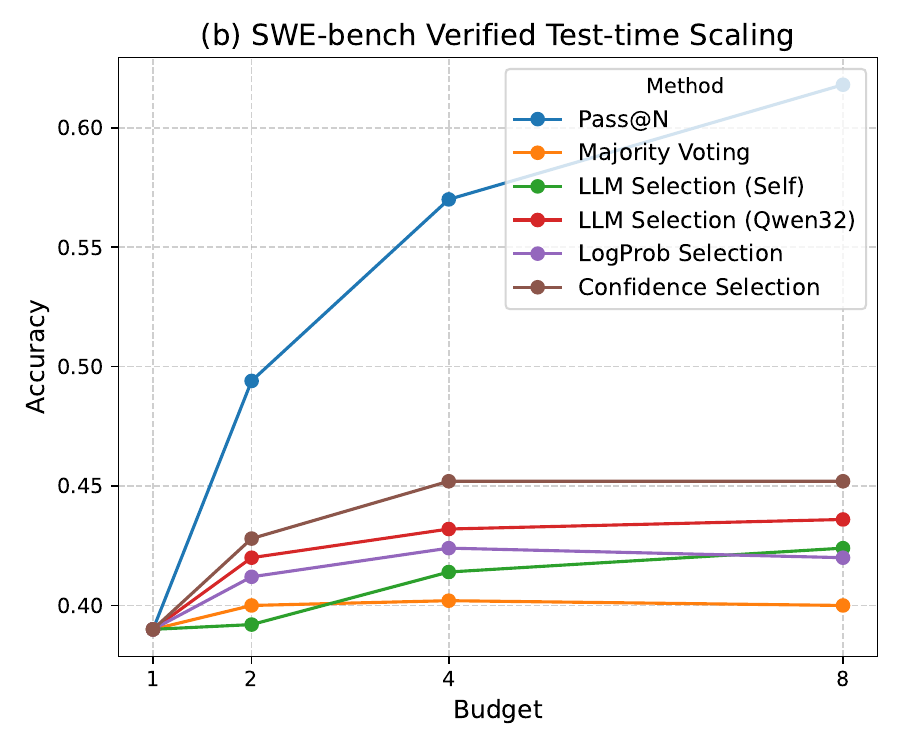}
    \end{minipage}
    \caption{Test-time scaling performance on BFCL V3 (a) and SWE-bench Verified (b).}
    \label{fig:tts}
\end{figure}

For reference, we report pass@N, which serves as an upper bound on success rate assuming the correct answer appears within the top-N generated candidates. Figure~\ref{fig:tts} (a)(b) summarizes the results on BFCL v3 and SWE-bench verified, with the number of sampled candidates varying from 1 to 16 and 1 to 8, respectively.
As shown in Figure~\ref{fig:tts}, pass@N demonstrates a clear upward trend as the number of candidates increases, confirming that parallel sampling substantially improves solution coverage. 
However, despite this clear improvement, most verifier-based approaches show limited or inconsistent gains. %
LLM-based selection methods perform relatively well at pass@2, as illustrated in Figure~\ref{fig:tts} (a). However, when selecting from larger candidate pools, their performance declines due to the inherent limitations of the models. Even using a larger model such as Qwen32 does not yield substantial improvements in candidate selection.
Among all strategies, Agent Confidence Select demonstrates the most stable and significant improvement, reaching 0.7668 at sixteen candidates on BFCL V3 and 0.452 at eight candidates on SWE-bench Verified. This suggests that internal confidence estimation offers a dependable signal for test-time selection, but even with this method, a noticeable gap remains between pass@N and the best verifier-based accuracy.

Compared to conventional test-time scaling, agentic TTS introduces a distinct perspective, as the trajectories naturally incorporate environment feedback. This feedback expands the design space of TTS, enabling the verifier to leverage not only model-internal signals but also external interaction outcomes. To explore this potential, we experimented with strategies that select candidate trajectories based on the log probability and confidence of the model’s responses to environment feedback. 
Specifically, the performance on multi-turn categories of BFCL v3 benchmark using Confidence Voting over environment-feedback tokens reached 0.4925, whereas applying it to agent-response tokens achieved 0.5200.
The results did not surpass expectations, suggesting that while environment-aware selection adds richer signals, it does not yet translate into substantial performance gains.

Overall, these findings highlight that while increasing candidate diversity improves solution coverage, the true performance gains are still constrained by the effectiveness of the verification strategy. Developing more robust, tool-augmented verifiers thus represents a crucial step toward realizing the full potential of test-time scaling in agentic LLMs.

\section{Conclusion and Future Work}
\label{sec:conclusion}
In this work, we presented Klear-AgentForge, a native agentic model designed to handle both tool-use and
coding tasks with the interaction with various environments. The model is trained through a pipeline that combines supervised fine-tuning, multi-turn reinforcement learning, model merge,
and test-time scaling. Our experiments
show that Klear-AgentForge-8B
attains balanced and competitive performance across different benchmarks within a single
unified model. The goal of this project is to fully invoke multi-domain agentic abilities through scaling the compute in posttraining, and thus we will continue
to explore the following directions:
\begin{itemize}
    \item Mid-training for improving agentic capabilities: Transforming a pretrained LLM, especially a small one, into a high-performing and robust agentic LLM is non-trivial. Supervised fine-tuning, when applied directly, risks causing abrupt distributional shift from our current experience, hindering the development of generalizable agentic skills. Therefore, effective midtraining acts as a distributional bridge,  smoothly transitioning a pretrained base model to the target agentic one. However, current mid-training mixtures are primarily composed of general domains like code, math, and instruction-following. To better support emerging agentic capabilities, we must explan its domains to include tasks that directly foster planning, tool-use, and data that contain interactions with different environments.
    \item Long and broad RL training: The key to scaling agentic RL lies in both increasing both the number of training steps and rollouts per example. However, the continuous environmental interaction required by agentic RL necessitates the development of specialized, efficient scaling methods, distinct from those used for single-turn reasoning tasks like RLVR. Furthermore, the development of lightweight modules capable of managing long context, particularly through an effective balance of global and local memory, is crucial for long-horizon agentic tasks.
    \item Small agentic LLM: Although larger models demonstrate stronger performance in our experiments, we contend that small language models (such as 8B size) offer a more compelling path for agentic AI. As~\cite{belcak2025small} suggests, their sufficient capability, inherent suitability, and superior economy for high-volume invocations make them the future of the field. From our current experience, it is still hard to train a single, unified small language model capable of excelling across all agentic tasks. A more promising approach may be to train a collection of specialized models, each optimized for a specific task or domain, and then employ a meta-agent to intelligently invoke this ensemble.
\end{itemize}

\section{Contributor List}
\label{sec:conclusion}
The authors are listed in order of the significance of their contributions, with those in project leadership roles appearing last.\\

\noindent
Qi Wang, Hongzhi Zhang, Jia Fu,  Kai Fu, Yahui Liu, Tinghai Zhang, Chenxi Sun, Gangwei Jiang, Jingyi Tang, Xingguang Ji, Yang Yue, Jingyuan Zhang, Fuzheng Zhang, Kun Gai, Guorui Zhou.  

\noindent
Corresponding authors: Qi Wang, Guorui Zhou

\setcitestyle{numbers}
\bibliography{main}
\setcitestyle{authoryear}

\end{CJK*}
\end{document}